\documentclass[10pt,twocolumn,letterpaper]{article}

\usepackage{amsfonts}
\usepackage{amssymb}
\usepackage{multirow}
\usepackage[pagenumbers]{cvpr} 

\usepackage[dvipsnames]{xcolor}

\definecolor{cvprblue}{rgb}{0.21,0.49,0.74}
\usepackage[pagebackref,breaklinks,colorlinks,citecolor=cvprblue]{hyperref}

\def\paperID{5815} 
\def\confName{CVPR}
\def\confYear{2024}

\title{\emph{RealCustom}: Narrowing Real Text Word for Real-Time Open-Domain Text-to-Image Customization}

\author{
Mengqi Huang\textsuperscript{\rm 1}\thanks{Works done during the intership at ByteDance.}, 
Zhendong Mao\textsuperscript{\rm 1}\thanks{Zhendong Mao is the corresponding author.}, 
Mingcong Liu\textsuperscript{\rm 2}, 
Qian He\textsuperscript{\rm 2}, 
Yongdong Zhang\textsuperscript{\rm 1} \\
\textsuperscript{\rm 1} University of Science and Technology of China;
\textsuperscript{\rm 2}ByteDance Inc. \\
{\tt\small \{huangmq\}@mail.ustc.edu.cn, \{zdmao, zhyd73\}@ustc.edu.cn, \{liumingcong, heqian\}@bytedance.com}
}

\begin{document}
\maketitle
\begin{abstract}

Text-to-image customization, which aims to synthesize text-driven images for the given subjects, has recently revolutionized content creation. Existing works follow the pseudo-word paradigm, i.e., represent the given subjects as pseudo-words and then compose them with the given text. However, the inherent entangled influence scope of pseudo-words with the given text results in a dual-optimum paradox, i.e., the similarity of the given subjects and the controllability of the given text could not be optimal simultaneously. We present \textbf{RealCustom} that, for the first time, disentangles similarity from controllability by precisely limiting subject influence to relevant parts only, achieved by gradually narrowing \textbf{real} text word from its general connotation to the specific subject and using its cross-attention to distinguish relevance. Specifically, RealCustom introduces a novel ``train-inference" decoupled framework: (1) during training, RealCustom learns general alignment between visual conditions to original textual conditions by a novel adaptive scoring module to adaptively modulate influence quantity; (2) during inference, a novel adaptive mask guidance strategy is proposed to iteratively update the influence scope and influence quantity of the given subjects to gradually narrow the generation of the real text word. Comprehensive experiments demonstrate the superior real-time customization ability of RealCustom in the open domain, achieving both unprecedented similarity of the given subjects and controllability of the given text for the first time. The project page is \url{https://corleone-huang.github.io/realcustom/}.

\end{abstract}    
\section{Introduction}
\label{sec:intro}

\begin{figure}
  \centering
  \includegraphics[width=1.0\linewidth]{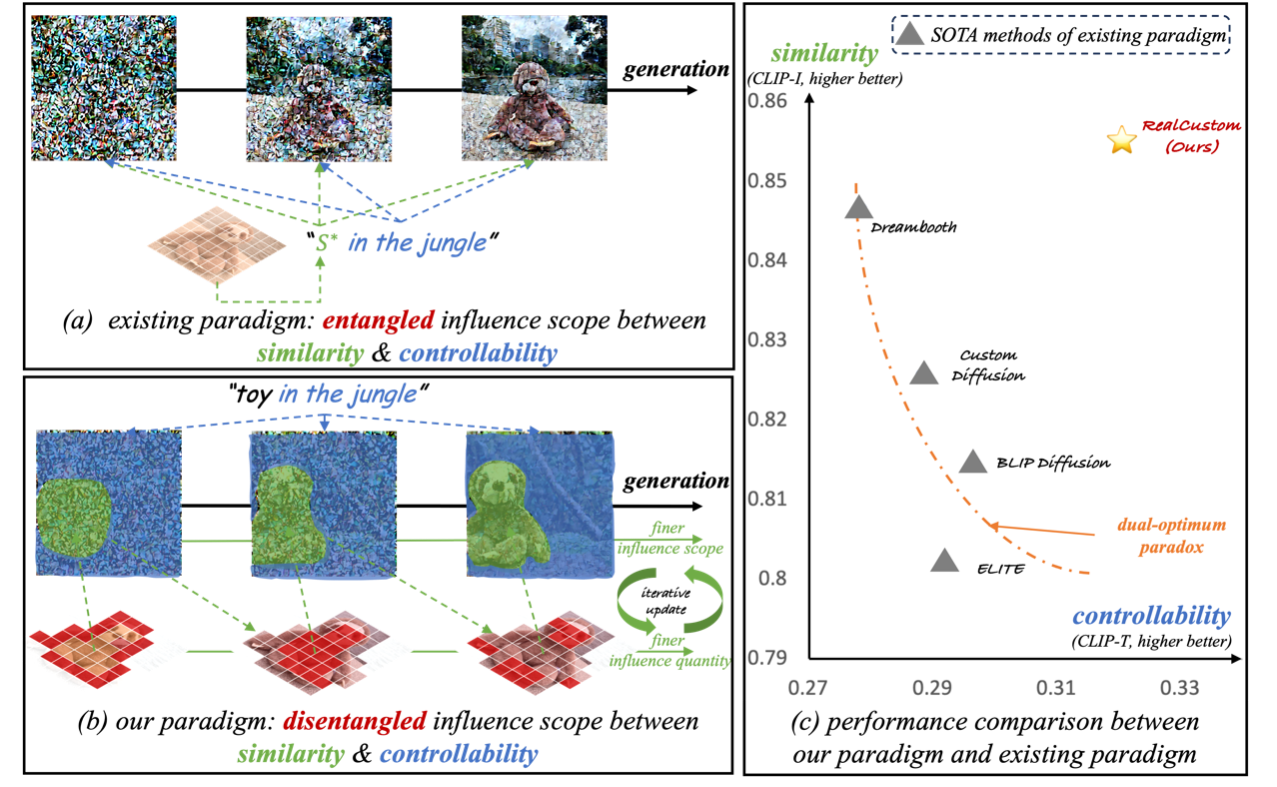}
  \caption{Comparison between the existing paradigm and ours. \textbf{(a)} The existing paradigm represents the \emph{given subject} as pseudo-words (\eg, $S^*$), which has entangled the same entire influence scope with the \emph{given text}, resulting in the \emph{dual-optimum paradox}, \ie, the similarity for the \emph{given subject} and the controllability for the \emph{given text} could not achieve optimum simultaneously. \textbf{(b)} We propose \emph{RealCustom}, a novel paradigm that, for the first time disentangles similarity from controllability by precisely limiting the \emph{given subjects} to influence only the relevant parts while the rest parts are purely controlled by the \emph{given text}. This is achieved by iteratively updating the influence scope and influence quantity of the \emph{given subjects}. \textbf{(c)} The quantitative comparison shows that our paradigm achieves both superior similarity and controllability than the state-of-the-arts of the existing paradigm. CLIP-image score (CLIP-I) and CLIP-text score (CLIP-T) are used to evaluate similarity and controllability. Refer to the experiments for details.}
  \label{intro}
\end{figure}

Recent significant advances in the customization of pre-trained large-scale text-to-image models \cite{ramesh2022hierarchical,rombach2022high,saharia2022photorealistic,chen2022re} (\ie, \emph{text-to-image customization}) has revolutionized content creation, receiving rapidly growing research interest from both academia and industry. This task empowers pre-trained models with the ability to generate imaginative text-driven scenes for subjects specified by users (\eg, a person's closest friends or favorite paintings), which is a foundation for AI-generated content (AIGC) and real-world applications such as personal image$\&$video creation \cite{chen2023dreamidentity}. The primary goal of customization is dual-faceted: (1) high-quality \emph{similarity}, \ie, the target subjects in the generated images should closely mirror the \emph{given subjects}; (2) high-quality \emph{controllability}, \ie, the remaining subject-irrelevant parts should consistently adhere to the control of the \emph{given text}.

Existing literature follows the \emph{pseudo-word} paradigm, \ie, (1) learning pseudo-words (\eg, $S^*$ \cite{gal2022image} or rare-tokens \cite{ruiz2023dreambooth}) to represent the given subjects; (2) composing these pseudo-words with the given text for the customized generation. Recent studies have focused on learning more comprehensive pseudo-words \cite{alaluf2023neural, zhang2023prospect, voynov2023p+, daras2022multiresolution, liu2023cones} to capture more subject information, \eg, different pseudo-words for different diffusion timesteps \cite{alaluf2023neural, zhang2023prospect} or layers \cite{voynov2023p+}. Meanwhile, others propose to speed up pseudo-word learning by training an encoder \cite{wei2023elite, li2023blip, shi2023instantbooth, gal2023designing} on object-datasets \cite{kuznetsova2020open}. In parallel, based on the learned pseudo-words, many works further finetune the pre-trained models \cite{kumari2023multi, ruiz2023dreambooth, wei2023elite, li2023blip} or add additional adapters \cite{shi2023instantbooth} for higher similarity. As more information of the given subjects is introduced into pre-trained models, the risk of overfitting increases, leading to the degradation of controllability. Therefore, various regularizations (\eg, $l_1$ penalty \cite{gal2022image, kumari2023multi, wei2023elite}, prior-preservation loss \cite{ruiz2023dreambooth}) are used to maintain controllability, which in turn sacrifices similarity. \emph{Essentially}, existing methods are trapped in a \emph{dual-optimum paradox}, \ie, the similarity and controllability can not be optimal simultaneously.

\begin{figure}
  \centering
  \includegraphics[width=1.0\linewidth]{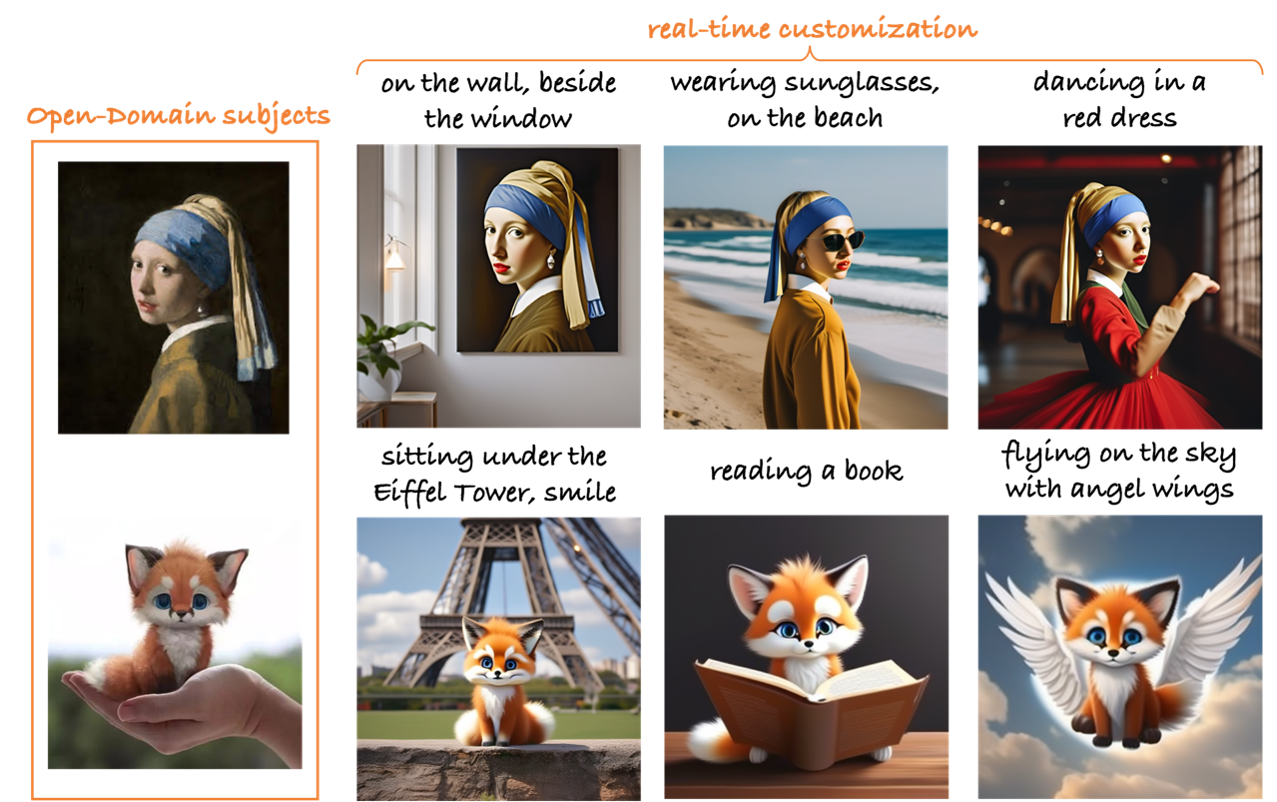}
  \caption{Generated customization results of our proposed novel paradigm \textbf{\emph{RealCustom}}. Given a \emph{single} image representing the given subject in the open domain (\textbf{\emph{any subjects}}, portrait painting, favorite toys, \etc), \emph{RealCustom} could generate realistic images that consistently adhere to the given text for the given subjects in real-time (\textbf{\emph{without any test-time optimization steps}}). }
  \label{intro2}
\end{figure}

We argue that the fundamental cause of this \emph{dual-optimum paradox} is rooted in the existing pseudo-word paradigm, where the similarity component (\ie, the pseudo-words) to generate the given subjects is intrinsically \emph{entangled} with the controllability component (\ie, the given text) to generate subject-irrelevant parts, causing an overall conflict in the generation, as illustrated in Fig. \ref{intro}(a). Specifically, this entanglement is manifested in the same entire influence scope of these two components. \ie, both the pseudo-words and the given text affect all generation regions. This is because each region is updated as a weighted sum of all word features through built-in textual cross-attention in pre-trained text-to-image diffusion models. Therefore, increasing the influence of the similarity component will simultaneously strengthen the similarity in the subject-relevant parts and weaken the influence of the given text in other irrelevant ones, causing the degradation of controllability, and \emph{vice versa}. Moreover, the necessary correspondence between pseudo-words and subjects confines existing methods to either lengthy test-time optimization \cite{gal2022image, ruiz2023dreambooth, kumari2023multi} or training \cite{li2023blip, wei2023elite} on object-datasets \cite{kuznetsova2020open} that have limited categories. As a result, the existing paradigm inherently has poor generalization capability for real-time open-domain scenarios in the real world.

In this paper, we present \textbf{\emph{RealCustom}}, a novel customization paradigm that, for the first time, disentangles the similarity component from the controllability component by precisely limiting the given subjects to influence only the relevant parts while maintaining other irreverent ones purely controlled by the given texts, achieving both high-quality similarity and controllability in a real-time open-domain scenario, as shown in Fig. \ref{intro2}. The core idea of \emph{RealCustom} is that, instead of representing subjects as pseudo-words, we could progressively narrow down the \textbf{\emph{real}} text words (\eg, ``toy") from their initial general connotation (\eg, various kinds o toys) to the specific subjects (\eg, the unique sloth toy), wherein the superior text-image alignment in pre-trained models' cross-attention can be leveraged to distinguish subject relevance, as illustrated in Fig. \ref{intro}(b). Specifically, at each generation step, (1) the influence scope of the given subject is identified by the target real word's cross-attention, with a higher attention score indicating greater relevance; (2) this influence scope then determines the influence quantity of the given subject at the current step, \ie, the amount of subject information to be infused into this scope; (3) this influence quantity, in turn, shapes a more accurate influence scope for the next step, as each step's generation result is based on the output of the previous. Through this iterative updating, the generation result of the real word is smoothly and accurately transformed into the given subject, while other irrelevant parts are completely controlled by the given text. 

Technically, \emph{RealCustom} introduces an innovative ``train-inference" decoupled framework: (1) During training, \emph{RealCustom} only learns the generalized alignment capabilities between visual conditions and pre-trained models' original text conditions on large-scale text-image datasets through a novel \emph{adaptive scoring module}, which modulates the influence quantity based on text and currently generated features. (2) During inference, real-time customization is achieved by a novel \emph{adaptive mask guidance strategy}, which gradually narrows down a real text word based on the learned alignment capabilities. Specifically, (1) the \emph{adaptive scoring module} first estimates the visual features' correlation scores with the text features and currently generated features, respectively. Then a timestep-aware schedule is applied to fuse these two scores. A subset of key visual features, chosen based on the fused score, is incorporated into pre-trained diffusion models by extending its textual cross-attention with another visual cross-attention. (2) The \emph{adaptive mask guidance strategy} consists of a \emph{text-to-image (T2I)} branch (with the visual condition set to $\boldsymbol{0}$) and a \emph{text$\&$image-to-image (TI2I)} branch (with the visual condition set to the given subject). Firstly, all layers' cross-attention maps of the target real word in the T2I branch are aggregated into a single one, selecting only high-attention regions as the influence scope. Secondly, in the TI2I branch, the influence scope is multiplied by currently generated features to produce the influence quantity and concurrently multiplied by the outputs of the visual cross-attention to avoid influencing subject-irrelevant parts.

Our contributions are summarized as follows:

\textcolor{blue}{Concepts.} For the first time, we (1) point out the \emph{dual-optimum paradox} is rooted in the existing pseudo-word paradigm's entangled influence scope between the similarity (\ie, pseudo-words representing the given subjects) and controllability (\ie, the given texts); (2) present \emph{RealCustom}, a novel paradigm that achieves disentanglement by gradually narrowing down \emph{real} words into the given subjects, wherein the given subjects' influence scope is limited based on the cross-attention of the real words.

\textcolor{blue}{Technology.} The proposed \emph{RealCustom} introduces a novel ``train-inference" decoupled framework: (1) during training, learning generalized alignment between visual conditions to original text conditions by the \emph{adaptive scoring module} to modulate influence quantity; (2) during inference, the \emph{adaptive mask guidance strategy} is proposed to narrow down a real word by iterative updating the given subject's influence scope and quantity.

\textcolor{blue}{Significance.} For the first time, we achieve (1) superior similarity and controllability \emph{simultaneously}, as shown in Fig. \ref{intro}(c); (2) real-time open-domain customization ability.


\section{Related Works}
\label{sec:related_works}

\subsection{Text-to-Image Customization}
Existing customization methods follow the \emph{pseudo-words} paradigm, \ie, representing the given subjects as \emph{pseudo-words} and then composing them with the given text for customization. Since the necessary correspondence between the pseudo-words and the given subjects, existing works are confined to either cumbersome test-time optimization-based \cite{gal2022image, ruiz2023dreambooth, kumari2023multi, alaluf2023neural, voynov2023p+, daras2022multiresolution, liu2023cones, dong2022dreamartist} or encoder-based \cite{wei2023elite, li2023blip, shi2023instantbooth, gal2023designing, jia2023taming, chen2023dreamidentity} that trained on object-datasets with limited categories. For example, in the optimization-based stream, DreamBooth \cite{ruiz2023dreambooth} uses a rare-token as the pseudo-word and further fine-tunes the entire pre-trained diffusion model for better similarity. Custom Diffusion \cite{kumari2023multi} instead finds a subset of key parameters and only optimizes them. The main drawback of this stream is that it requires lengthy optimization times for each new subject. As for the encoder-based stream, the recent ELITE \cite{wei2023elite} uses a local mapping network to improve similarity, while BLIP-Diffusion \cite{li2023blip} introduces a multimodal encoder for better subject representation. These encoder-based works usually show less similarity than optimization-based works and generalize poorly to unseen categories in training. \emph{In summary}, the entangled influence scope of pseudo-words and the given text naturally limits the current works from achieving both optimal similarity and controllability, as well as hindering real-time open-domain customization.

\subsection{Cross-Attention in Diffusion Models}

Text guidance in modern large-scale text-to-image diffusion models \cite{ramesh2022hierarchical,rombach2022high,saharia2022photorealistic,chen2022re,balaji2022ediffi} is generally performed using the cross-attention mechanism. Therefore, many works propose to manipulate the cross-attention map for text-driven editing \cite{hertz2022prompt, cao2023masactrl} on generated images or real images via inversion \cite{song2020denoising}, \eg, Prompt-to-Prompt \cite{hertz2022prompt} proposes to reassign the cross-attention weight to edit the generated image. Another branch of work focuses on improving cross-attention either by adding additional spatial control \cite{li2023gligen, li2023guiding} or post-processing to improve semantic alignment \cite{chefer2023attend, li2023divide}. Meanwhile, a number of works \cite{wu2023diffumask, xiao2023text, wang2023diffusion} propose using cross-attention in diffusion models for discriminative tasks such as segmentation. However, different from the existing literature, the core idea of \emph{RealCustom} is to gradually narrow a real text word from its initial general connotation (\eg, whose cross-attention could represent any toy with various types of shapes and details) to the unique given subject (\eg, whose cross-attention accurately represents the unique toy), which is completely unexplored.

\begin{figure*}
  \centering
  \includegraphics[width=1.0\linewidth]{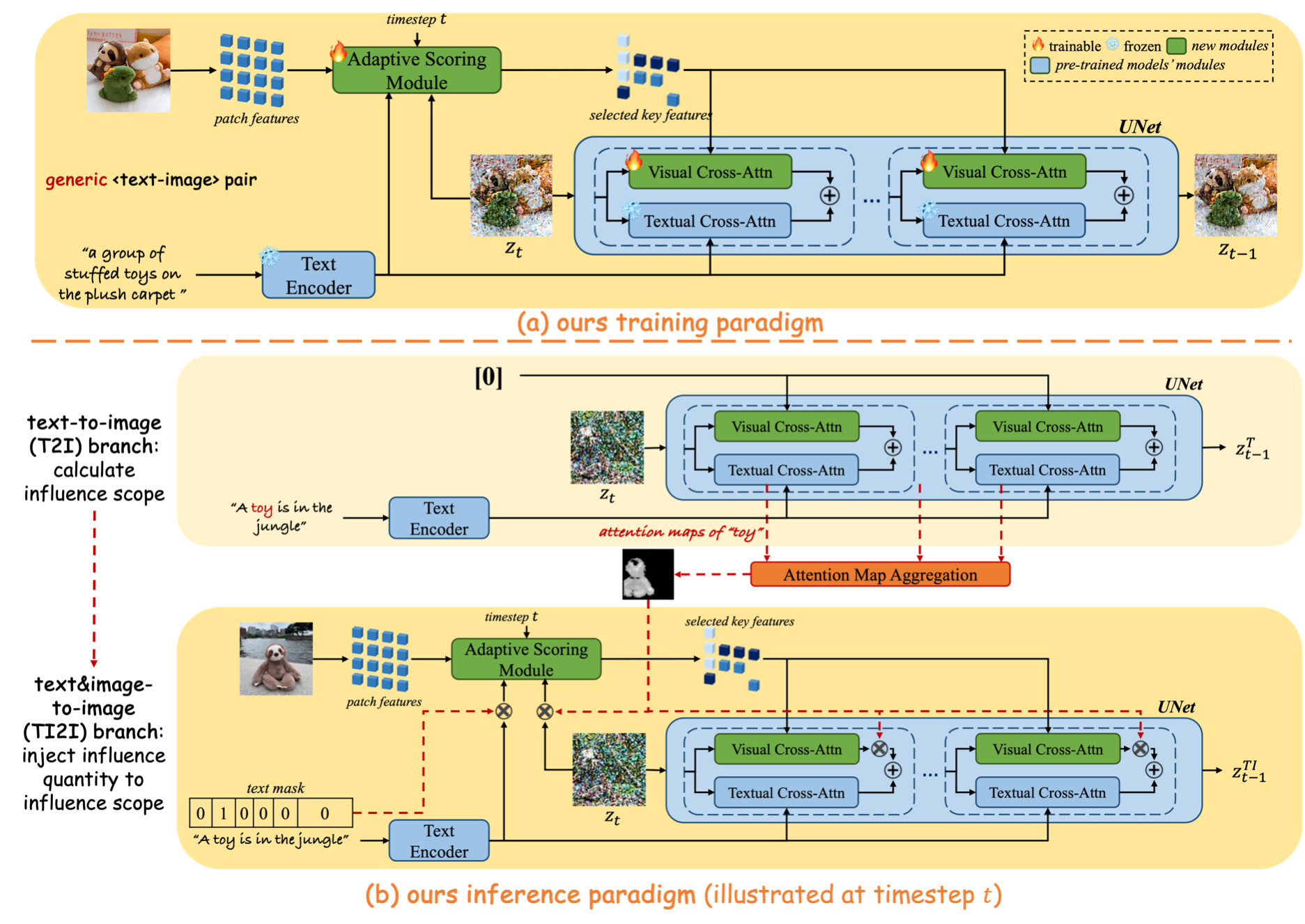}
  \caption{Illustration of our proposed \emph{RealCustom}, which employs a novel ``train-inference" decoupled framework: (a) During training, general alignment between visual and original text conditions is learned by the proposed \emph{adaptive scoring module}, which accurately derives visual conditions based on text and currently generated features. (b) During inference, progressively narrowing down a real word (\eg, ``toy") from its initial general connotation to the given subject (\eg, the unique brown sloth toy) by the proposed \emph{adaptive mask guidance strategy}, which consists of two branches, \ie, a text-to-image (T2I) branch where the visual condition is set to $\boldsymbol{0}$, and a text$\&$image-to-image (TI2I) branch where the visual condition is set to the given subject. The T2I branch aims to calculate the influence scope by aggregating the target real word's (\eg, ``toy") cross-attention, while the TI2I branch aims to inject the influence quantity into this scope.}
  \label{framework}
\end{figure*}

\section{Methodology}
\label{sec:method}

In this study, we focus on the most general customization scenario: with only a \emph{single} image representing the given subject, generating new high-quality images for that subject from the given text. The generated subject may vary in location, pose, style, \etc, yet it should maintain high \emph{similarity} with the given one. The remaining parts should consistently adhere to the given text, thus ensuring \emph{controllability}.


The proposed \emph{RealCustom} introduces a novel ``train-inference" decoupled paradigm as illustrated in Fig. \ref{framework}. Specifically, during training, \emph{RealCustom} learns general alignment between visual conditions and the original text conditions of pre-trained models. During inference, based on the learned alignment capability, \emph{RealCustom} gradually narrow down the generation of the real text words (\eg, ``toy") into the given subject (\eg, the unique brown sloth toy) by iterative updating each step's influence scope and influence quantity of the given subject.


We first briefly introduce the preliminaries in Sec. \ref{preliminaries}. The training and inference paradigm of \emph{RealCustom} will be elaborated in detail in Sec. \ref{paradigm_train} and Sec. \ref{paradigm_inference}, respectively. 


\subsection{Preliminaries}
\label{preliminaries}

Our paradigm is implemented over Stable Diffusion \cite{rombach2022high}, which consists of two components, \ie, an autoencoder and a conditional UNet \cite{ronneberger2015u} denoiser. Firstly, given an image $\boldsymbol{x} \in \mathbb{R}^{H \times W \times 3}$, the encoder $\mathcal{E(\cdot)}$ of the autoencoder maps it into a lower dimensional latent space as $\boldsymbol{z}=\mathcal{E}(\boldsymbol{x}) \in \mathbb{R}^{h \times w \times c}$, where $f=\frac{H_0}{h} = \frac{W_0}{w}$ is the downsampling factor and $c$ stands for the latent channel dimension. The corresponding decoder $\mathcal{D}(\cdot)$ maps the latent vectors back to the image as $\mathcal{D}(\mathcal{E}(\boldsymbol{x})) \approx \boldsymbol{x}$. Secondly, the conditional denoiser $\epsilon_\theta (\cdot)$ is trained on this latent space to generate latent vectors based on the text condition $y$. The pre-trained CLIP text encoder \cite{radford2021learning} $\tau_{\text{text}}(\cdot)$ is used to encode the text condition $y$ into text features $\boldsymbol{f_{ct}} = \tau_{\text{text}}(y)$. Then, the denoiser is trained with mean-squared loss:
\begin{equation}
    L:=\mathbb{E}_{\boldsymbol{z} \sim \mathcal{E}(\boldsymbol{x}), \boldsymbol{f_y}, \boldsymbol{\epsilon} \sim \mathcal{N}(\boldsymbol{0},\boldsymbol{\text{I}}), t}\left[\left\|\boldsymbol{\epsilon} -\epsilon_\theta\left(\boldsymbol{z_t}, t, \boldsymbol{f_{ct}}\right)\right\|_2^2\right],
\end{equation}
where $\boldsymbol{\epsilon}$ denotes for the unscaled noise and $t$ is the timestep. $\boldsymbol{z_t}$ is the latent vector that noised according to $t$:
\begin{equation}
    \boldsymbol{z_t} = \sqrt{\hat{\alpha}_t} \boldsymbol{z_0} + \sqrt{1 - \hat{\alpha}_t} \boldsymbol{\epsilon},
\end{equation}
where $\hat{\alpha}_t \in [0,1]$ is the hyper-parameter that modulates the quantity of noise added. Larger $t$ means smaller $\hat{\alpha}_t$ and thereby a more noised latent vector $\boldsymbol{z_t}$. During inference, a random Gaussian noise $\boldsymbol{z_T}$ is iteratively denoised to $\boldsymbol{z_0}$, and the final generated image is obtained through $\boldsymbol{x^{'}}=\mathcal{D}(\boldsymbol{z_0})$.

The incorporation of text condition in Stable Diffusion is implemented as textual cross-attention:
\begin{equation}
    \label{attention}
    \text{Attention}(\boldsymbol{Q},\boldsymbol{K},\boldsymbol{V}) = \text{Softmax}(\frac{\boldsymbol{Q}\boldsymbol{K}^{\top}}{\sqrt{d}})\boldsymbol{V},
\end{equation}
where the query $\boldsymbol{Q} = \boldsymbol{W_Q} \cdot \boldsymbol{f_i}$, key $\boldsymbol{K} = \boldsymbol{W_K} \cdot \boldsymbol{f_{ct}}$ and value $\boldsymbol{V} = \boldsymbol{W_V} \cdot \boldsymbol{f_{ct}}$. $\boldsymbol{W_Q}, \boldsymbol{W_K}, \boldsymbol{W_V}$ are weight parameters of query, key and value projection layers. $\boldsymbol{f_i}, \boldsymbol{f_{ct}}$ are the latent image features and text features, and $d$ is the channel dimension of key and query features. The latent image feature is then updated with the attention block output.

\subsection{Training Paradigm}
\label{paradigm_train}

As depicted in Fig. \ref{framework}(a), the text $y$ and image $x$ are first encoded into text features $\boldsymbol{f_{ct}} \in \mathbb{R}^{n_t \times c_t}$ and image features $\boldsymbol{f_{ci}} \in \mathbb{R}^{n_i \times c_i}$ by the pre-trained CLIP text/image encoders \cite{radford2021learning} respectively. Here, $n_t, c_t, n_i, c_i$ are text feature number/dimension and image feature number/dimension, respectively. Afterward, the \emph{adaptive scoring module} takes the text features $\boldsymbol{f_{ct}}$, currently generated features $\boldsymbol{z_t} \in \mathbb{R}^{h \times w \times c}$, and timestep $t$ as inputs to estimate the score for each features in $\boldsymbol{f_{ci}}$, selecting a subset of key ones as the visual condition $\boldsymbol{\hat{f}_{ci}} \in \mathbb{R}^{\hat{n}_i \times c_i}$, where $\hat{n}_i < n_i$ is the selected image feature number. Next, we extend textual cross-attention with another visual cross-attention to incorporate the visual condition $\boldsymbol{\hat{f}_{yi}}$. Specifically, Eq. \ref{attention} is rewritten as:
\begin{multline}
    \label{dual_attention}
    \text{Attention}(\boldsymbol{Q},\boldsymbol{K},\boldsymbol{V},\boldsymbol{K_i},\boldsymbol{V_i}) = \\
    \text{Softmax}(\frac{\boldsymbol{Q}\boldsymbol{K}^{\top}}{\sqrt{d}})\boldsymbol{V} + \text{Softmax}(\frac{\boldsymbol{Q}\boldsymbol{K_i}^{\top}}{\sqrt{d}})\boldsymbol{V_i},
\end{multline}
where the new key $\boldsymbol{K_i} = \boldsymbol{W_{Ki}} \cdot \boldsymbol{\hat{f}_{ci}}$, value $\boldsymbol{V_i} = \boldsymbol{W_{Vi}} \cdot \boldsymbol{\hat{f}_{ci}}$ are added. $\boldsymbol{W_{Ki}}$ and $\boldsymbol{W_{Vi}}$ are weight parameters. During training, only the \emph{adaptive scoring module} and projection layers $\boldsymbol{W_{Ki}}, \boldsymbol{W_{Vi}}$ in each attention block are trainable, while other pre-trained models' weight remains frozen.


\textbf{Adaptive Scoring Module}. On the one hand, the generation of the diffusion model itself, by nature, is a coarse-to-fine process with noise removed and details added step by step. In this process, different steps focus on different degrees of subject detail \cite{balaji2022ediffi}, spanning from global structures in the early to local textures in the latter. Accordingly, the importance of each image feature also dynamically changes. To smoothly narrow the real text word, the image condition of the subject should also adapt synchronously, providing guidance from coarse to fine grain. This requires equipping \emph{RealCustom} with the ability to estimate the importance score of different image features. On the other hand, utilizing all image features as visual conditions results in a ``train-inference'' gap. This arises because, unlike the training stage, where the same images as the visual conditions and inputs to the denoiser $\epsilon_\theta$, the given subjects, and the inference generation results should maintain similarity only in the subject part. Therefore, this gap can degrade both similarity and controllability in inference.

\begin{figure}
  \centering
  \includegraphics[width=1.0\linewidth]{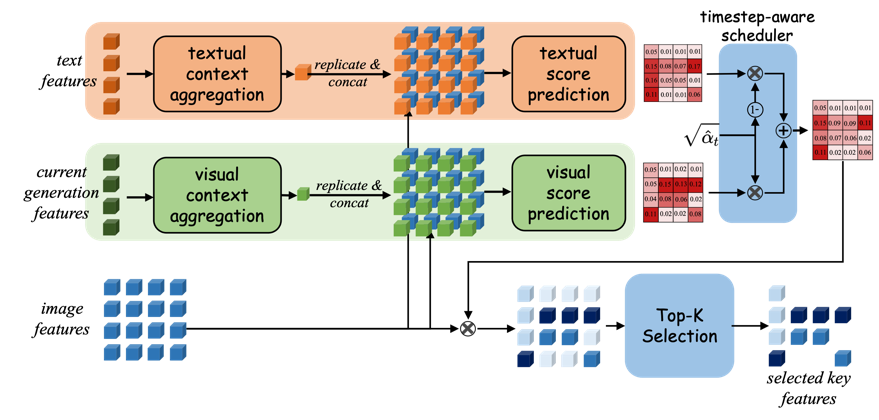}
  \caption{Illustration of \emph{adaptive scoring module}. Text features and currently generated features are first aggregated into the textual and visual context, which are then spatially concatenated with image features to predict textual and visual scores. These scores are then fused based on the current timestep. Ultimately, only a subset of the key features is selected based on the fused score.}
  \label{score_module}
\end{figure}

The above rationale motivates the \emph{adaptive scoring module}, which provides smooth and accurate visual conditions for customization. As illustrated in Fig. \ref{score_module}, the text $\boldsymbol{f_{ct}} \in \mathbb{R}^{n_t \times c_t}$ and currently generated features $\boldsymbol{z_t} \in \mathbb{R}^{h \times w \times c}=\mathbb{R}^{n_z \times c}$ are first aggregated into the textual context $\boldsymbol{C_{\text{textual}}}$ and visual context $\boldsymbol{C_{\text{visual}}}$ through weighted pooling: 
\begin{align}
    \small
    \boldsymbol{A_{\text{textual}}} = \text{Softmax}(\boldsymbol{f_{ct}} \boldsymbol{W_a^t}) \in \mathbb{R}^{n_t \times 1} \\
    \small
    \boldsymbol{A_{\text{visual}}} = \text{Softmax}(\boldsymbol{z_t} \boldsymbol{W_a^v}) \in \mathbb{R}^{n_z \times 1} \\
    \small
    \boldsymbol{C_{\text{textual}}} = \boldsymbol{A_{\text{textual}}^{\top}} \boldsymbol{f_y} \in \mathbb{R}^{1 \times c_t},
    \boldsymbol{C_{\text{visual}}} = \boldsymbol{A_{\text{visual}}^{\top}} \boldsymbol{z_t} \in \mathbb{R}^{1 \times c},
\end{align}
where $\boldsymbol{W_a^t} \in \mathbb{R}^{c_t \times 1}, \boldsymbol{W_a^v} \in \mathbb{R}^{c \times 1}$ are weight parameters, and ``Softmax" is operated in the number dimension. These contexts are then spatially replicated and concatenated with image features $\boldsymbol{f_{ci}} \in \mathbb{R}^{n_i \times c_i}$ to estimate the textual score $\boldsymbol{S_{\text{textual}}} \in \mathbb{R}^{n_i \times 1}$ and visual score $\boldsymbol{S_{\text{visual}}} \in \mathbb{R}^{n_i \times 1}$ respectively. These two scores are predicted by two lightweight score-net, which are implemented as two-layer MLPs. 

Considering that the textual features are roughly accurate and the generated features are gradually refined, a timestep-aware schedule is proposed to fuse these two scores:
\begin{equation}
    \boldsymbol{S} = (1 - \sqrt{\hat{\alpha}_t}) \boldsymbol{S_{\text{textual}}} + \sqrt{\hat{\alpha}_t} \boldsymbol{S_{\text{visual}}},
\end{equation}
where $\sqrt{\hat{\alpha}_t}$ is the hyperparameter of pre-trained diffusion models that modulate the amount of noise added to generated features. Then a softmax activation is applied to the fused score since our focus is on highlighting the comparative significance of each image feature vis-à-vis its counterparts: $\boldsymbol{S} = \text{Softmax}(\boldsymbol{S})$. The fused scores are multiplied with the image features to enable the learning of score-nets:


\begin{equation}
    \boldsymbol{f_{ci}} = \boldsymbol{f_{ci}} \circ (1 + \boldsymbol{S}),
\end{equation}
where $\circ$ denotes the element-wise multiply. Finally, given a Top-K ratio $\gamma_{\text{num}} \in [0,1]$, a sub-set of key features with highest scores are selected as the output $\boldsymbol{\hat{f}_{yi}} \in \mathbb{R}^{\hat{n}_i \times c_i}$, where $\hat{n}_i = \gamma_{\text{num}} n_i$. To enable flexible inference with different $\gamma_{\text{num}}$ without performance degradation, we propose to use a uniformly random ratio during training:
\begin{equation}
    \gamma_{\text{num}} = \text{uniform}[\gamma_{\text{num}}^{\text{low}}, \gamma_{\text{num}}^{\text{high}}],
\end{equation}
where $\gamma_{\text{num}}^{\text{low}}, \gamma_{\text{num}}^{\text{high}}$ are set to $0.3, 1.0$, respectively.

\subsection{Inference Paradigm}
\label{paradigm_inference}

The inference paradigm of \emph{RealCustom} consists of two branches, \ie, a text-to-image (T2I) branch where the visual input is set to $\boldsymbol{0}$ and a text$\&$image-to-image (TI2I) branch where the visual input is set to given subjects, as illustrated in Fig. \ref{framework}(b). These two branches are connected by our proposed \emph{adaptive mask guidance strategy}. Specifically, given previous step's output $\boldsymbol{z_t}$, a pure text conditional denoising process is performed in T2I branch to get the output $\boldsymbol{z_{t-1}^{T}}$, where all layers cross-attention map of the target real word (\eg, ``toy") is extracted and resized to the same resolution (the same as the largest map size, \ie, $64 \times 64$ in Stable Diffusion). The aggregated attention map is denoted as $\boldsymbol{M} \in \mathbb{R}^{64 \times 64}$. Next, a Top-K selection is applied, \ie, given the target ratio $\gamma_{\text{scope}} \in [0,1]$, only $\gamma_{\text{scope}} \times 64 \times 64$ regions with the highest cross-attention score will remain, while the rest will be set to $0$. The selected cross-attention map $\boldsymbol{\bar{M}}$ is normalized by its maximum value as:
\begin{equation}
    \label{max_norm}
    \boldsymbol{\hat{M}} = \frac{\boldsymbol{\bar{M}}}{\text{max}(\boldsymbol{\bar{M})}},
\end{equation}
where $\text{max}(\cdot)$ represents the maximum value. The rationale behind this is that even in these selected parts, the subject relevance of different regions is also different. 



In the TI2I branch, the influence scope $\boldsymbol{\hat{M}}$ is first multiplied by currently generated feature $\boldsymbol{z_t}$ to provide accurate visual conditions for current generation step. The reason is that only subject-relevant parts should be considered for the calculation of influence quantity. Secondly, $\boldsymbol{\hat{M}}$ is multiplied by the visual cross-attention results to prevent negative impacts on the controllability of the given texts in other subject-irrelevant parts. Specifically, Eq. \ref{dual_attention} is rewritten as:
\begin{multline}
    \label{dual_attention_masked}
    \text{Attention}(\boldsymbol{Q},\boldsymbol{K},\boldsymbol{V},\boldsymbol{K_i},\boldsymbol{V_i}) = \\
    \text{Softmax}(\frac{\boldsymbol{Q}\boldsymbol{K}^{\top}}{\sqrt{d}})\boldsymbol{V} + (\text{Softmax}(\frac{\boldsymbol{Q}\boldsymbol{K_i}^{\top}}{\sqrt{d}})\boldsymbol{V_i}) \boldsymbol{\hat{M}},
\end{multline}
where the necessary resize operation is applied to match the size of $\boldsymbol{\hat{M}}$ with the resolution of each cross-attention block. The denoised output of TI2I branch is denoted as $\boldsymbol{z_{t-1}^{TI}}$. The classifer-free guidance \cite{ho2022classifier} is extended to produce next step's denoised latent feature $\boldsymbol{z_{t-1}}$ as:
\begin{equation}
    \boldsymbol{z_{t-1}} = \epsilon_\theta(\emptyset) + \omega_t ( \boldsymbol{z_{t-1}^{T}} - \epsilon_\theta(\emptyset)) + \omega_i (\boldsymbol{z_{t-1}^{TI}} - \boldsymbol{z_{t-1}^{T}}),
\end{equation}
where $\epsilon_\theta(\emptyset)$ is the unconditional denoised output. 

With the smooth and accurate influence quantity of the given subject injected into the current step, the generation of the real word will gradually be narrowed from its initial general connotation to the specific subject, which will shape a more precise influence scope for the generation of the next step. Through this iterative updating and generation, we achieve real-time customization where the similarity for the given subject is disentangled with the controllability for the given text, leading to an optimal of both. More importantly, since both the \emph{adaptive scoring module} as well as visual cross-attention layers are trained on general text-image datasets, the inference could be generally applied to any categories by using any target real words, enabling excellent open-domain customization capability.
\section{Experiments}
\label{sec:exps}

\begin{table*}
\begin{minipage}[p]{0.7\textwidth}
\centering
\resizebox{\linewidth}{!}{%
\begin{tabular}{ccccccc}
    \toprule
    \multirow{2}*{Methods} & \multicolumn{2}{c}{\textbf{\emph{controllability}}} & \multicolumn{2}{c}{\textbf{\emph{similarity}}} & \multicolumn{1}{c}{\textbf{\emph{efficiency}}} &  \\
    \cmidrule(lr){2-3} \cmidrule(lr){4-5} \cmidrule(lr){6-6}
         & CLIP-T $\uparrow$ & ImageReward $\uparrow$ & CLIP-I $\uparrow$ & DINO-I $\uparrow$ & \footnotesize{test-time optimize steps} \\
    \midrule
    Textual Inversion \cite{gal2022image} & 0.2546 & -0.9168 & 0.7603 & 0.5956 & 5000 \\
    DreamBooth \cite{ruiz2023dreambooth} & 0.2783 & 0.2393 & 0.8466 & 0.7851 & 800 \\
    Custom Diffusion \cite{kumari2023multi} & 0.2884 & 0.2558 & 0.8257 & 0.7093 & 500 \\
    \hline
    ELITE \cite{wei2023elite} & 0.2920 & 0.2690 & 0.8022 & 0.6489 & 0 (real-time)\\    
    BLIP-Diffusion \cite{li2023blip} & 0.2967 & 0.2172 & 0.8145 & 0.6486 & 0 (real-time) \\
    \hline
    \textbf{\emph{RealCustom}(ours)} & \textbf{0.3204} & \textbf{0.8703} & \textbf{0.8552} & \textbf{0.7865} & \textbf{0 (real-time)} \\
    \bottomrule
  \end{tabular}
}
\end{minipage}
\begin{minipage}[p]{0.3\textwidth} 
    \centering 
    \resizebox{\linewidth}{!}{%
    \includegraphics[width=60mm]{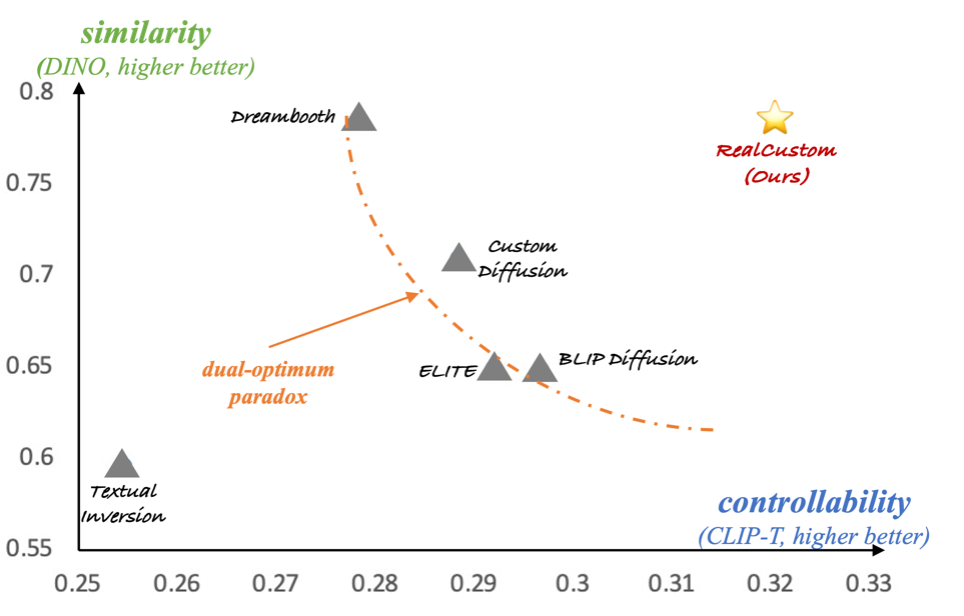} 
    }
\end{minipage}
\caption{Quantitative comparisons with existing methods. \textbf{Left}: Our proposed \emph{RealCustom} outperforms existing methods in all metrics, \ie, (1) for controllability, achieving 8.1\% and 223.5\% improvements on CLIP-T and ImageReward, respectively. The significant improvement on ImageReward also validates that \emph{RealCustom} could generate customized images with much higher quality (higher aesthetic score); (2) for similarity, we also achieve state-of-the-art performance on both CLIP-I and DINO-I. \textbf{Right}: We plot the ``CLIP-T verse DINO", showing that the existing methods are trapped into the \emph{dual-optimum paradox}, while \emph{RealCustom} completely get rid of it and achieve both high-quality similarity and controllability. The same conclusion in ``CLIP-T verse CLIP-I" can be found in Fig. \ref{intro}(c). } 
\label{compare_1} 
\end{table*}

\begin{figure*}
  \centering
  \includegraphics[width=1.0\linewidth]{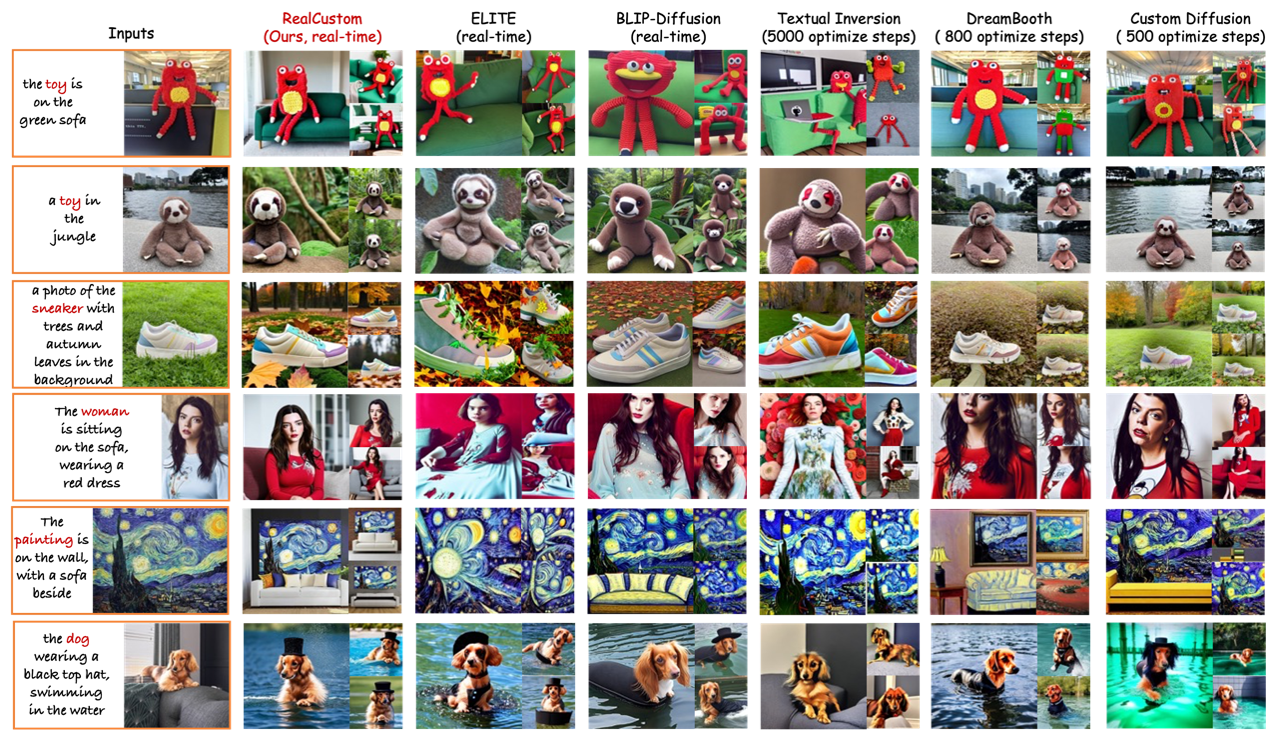}
  \caption{Qualitative comparison with existing methods. \emph{RealCustom} could produce much higher quality customization results that have better similarity with the given subject and better controllability with the given text compared to existing works. Moreover, \emph{RealCustom} shows superior diversity (different subject poses, locations, \etc) and generation quality (\eg, the ``autumn leaves" scene in the third row).}
  \label{visual_comparison}
\end{figure*}

\begin{figure*}
  \centering
  \includegraphics[width=1.0\linewidth]{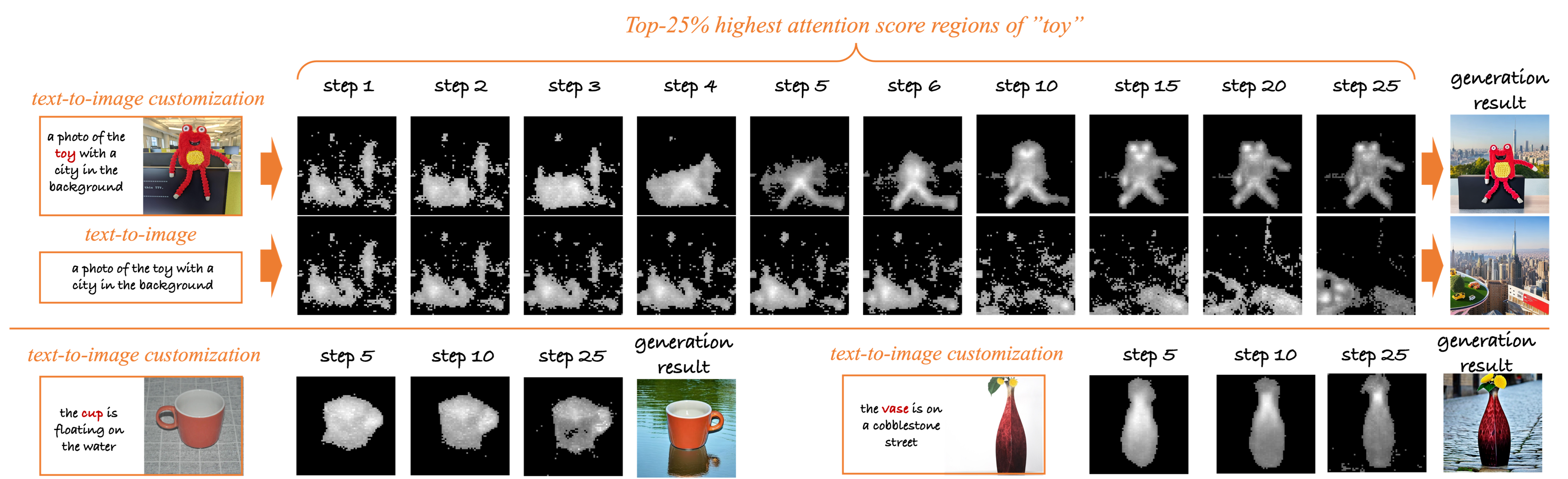}
  \caption{Illustration of gradually narrowing the real words into the given subjects. \textbf{Upper}: \emph{RealCustom} generated results (first row) and the original text-to-image generated result (second row) by pre-trained models with the same seed. The mask is visualized by the Top-25\% highest attention score regions of the real word ``toy". We could observe that starting from the same state (the same mask since there's no information of the given subject is introduced at the beginning), \emph{RealCustom} gradually forms the structure and details of the given subject by our proposed \emph{adaptive mask strategy}, achieving the open-domain zero-shot customization. \textbf{Lower}: More visualization cases. }
  \label{mask1}
\end{figure*}

\begin{table}
  \footnotesize
  \centering
  \begin{tabular}{ccc}
    \toprule
    inference setting & CLIP-T $\uparrow$ & CLIP-I $\uparrow$ \\
    \hline
    $\gamma_{\text{scope}} = 0.1$ & 0.32 & 0.8085 \\
    $\gamma_{\text{scope}} = 0.2$ & 0.3195 & 0.8431 \\
    $\gamma_{\text{scope}} = 0.25$ & \textbf{0.3204} & \textbf{0.8552} \\
    $\gamma_{\text{scope}} = 0.25$, binary & 0.294 & 0.8567 \\
    $\gamma_{\text{scope}} = 0.3$ & 0.3129 & 0.8578 \\
    $\gamma_{\text{scope}} = 0.4$ & 0.3023 & 0.8623 \\
    $\gamma_{\text{scope}} = 0.5$ & 0.285 & 0.8654 \\
    \bottomrule
  \end{tabular}
  \caption{Ablation of different $\gamma_{\text{scope}}$, which denotes the influence scope of the given subject in \emph{RealCustom} during inference. ``binary" means using binary masks instead of max norm in Eq. \ref{max_norm}. }
  \label{ablation_1}
\end{table}

\begin{figure}
  \centering
  \includegraphics[width=1.0\linewidth]{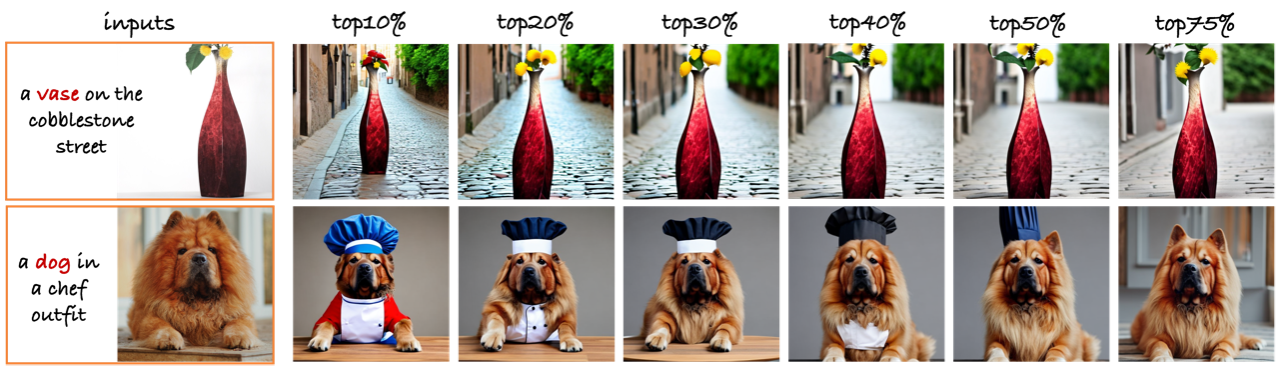}
  \caption{Visualization of different influence scope. }
  \label{topk_region}
\end{figure}

\subsection{Experimental Setups}

\hspace*{1em} \textbf{Implementation.} \emph{RealCustom} is implemented on Stable Diffusion and trained on the filtered subset of Laion-5B \cite{schuhmann2022laion} based on aesthetic score, using 16 A100 GPUs for 16w iterations with 1e-5 learning rate. Unless otherwise specified, DDIM sampler \cite{song2020denoising} with 50 sample steps is used for sampling and the classifier-free guidance $\omega_t, \omega_i$ is 7.5 and 12.5. Top-K ratios $\gamma_{\text{num}}=0.8$, $\gamma_{\text{scope}}=0.25$.

\textbf{Evaluation.} \textbf{\emph{Similarity.}} We use the state-of-the-art segmentation model (\ie, SAM \cite{kirillov2023segment}) to segment the subject, and then evaluate with both CLIP-I and DINO \cite{caron2021emerging} scores, which are average pairwise cosine similarity CLIP ViT-B/32 or DINO embeddings of the segmented subjects in generated and real images. \textbf{\emph{Controllability.}} We calculate the cosine similarity between prompt and image CLIP ViT-B/32 embeddings (CLIP-T). In addition, ImageReward \cite{xu2023imagereward} is used to evaluate controllability and aesthetics (quality).


\textbf{Prior SOTAs.} We compare with existing paradigm of both optimization-based (\ie, Textual Inversion\cite{gal2022image}, DreamBooth \cite{ruiz2023dreambooth}, CustomDiffusion \cite{kumari2023multi}) and encoder-based (ELITE\cite{wei2023elite}, BLIP-Diffusion\cite{li2023blip}) state-of-the-arts.

\subsection{Main Results}


\hspace*{1em} \textbf{Quantitative results.} As shown in Tab. \ref{compare_1}, \emph{RealCustom} outperforms existing methods in all metrics: (1) for controllability, we improve CLIP-T and ImageReward by 8.1\% and 223.5\%, respectively. The significant improvement in ImageReward shows that our paradigm generates much higher quality customization; (2) for similarity, we also achieve state-of-the-art performance on both CLIP-I and DINO-I. The figure of ``CLIP-T verse DINO" validates that the existing paradigm is trapped into the \emph{dual-optimum paradox}, while RealCustom effectively eradicates it.

\textbf{Qualitative results.} As shown in Fig. \ref{visual_comparison}, \emph{RealCustom} demonstrates superior zero-shot open-domain customization capability (\eg, the rare shaped toy in the first row), generating higher-quality custom images that have better similarity with the given subject and better controllability with the given text compared to existing works. 

\subsection{Ablations}

\textbf{Effectiveness of \emph{adaptive mask guidance strategy}.} We first visualize the narrowing down process of the real word by the proposed adaptive mask guidance strategy in Fig. \ref{mask1}. We could observe that starting from the same state (the same mask since there's no information of the given subject is introduced at the first step), \emph{RealCustom} gradually forms the structure and details of the given subject, achieving the open-domain zero-shot customization while remaining other subject-irrelevant parts (\eg, the city background) completely controlled by the given text. 

\begin{table}
  \centering
  \footnotesize
  \begin{tabular}{lccc}
    \toprule
    ID & settings & CLIP-T $\uparrow$ & CLIP-I $\uparrow$ \\
    \hline
    1 & full model, $\gamma_{\text{num}}=0.8$ & \textbf{0.3204} & \textbf{0.8552} \\
    2 & \emph{w/o} adaptive scoring module & 0.3002 & 0.8221 \\
    \hline 
    3 & textual score only, $\gamma_{\text{num}}=0.8$ & 0.313 & 0.8335 \\
    4 & visual score only, $\gamma_{\text{num}}=0.8$ & 0.2898 & 0.802 \\
    5 & (textual + visual) / 2, $\gamma_{\text{num}}=0.8$ & 0.3156 & 0.8302 \\
    \hline
    6 & full model, $\gamma_{\text{num}}=0.9$ & 0.315 & 0.8541 \\
    7 & full model, $\gamma_{\text{num}}=0.7$ & 0.3202 & 0.8307 \\
    \bottomrule
  \end{tabular}
  \caption{Ablation of the adaptive scoring module, where $\gamma_{\text{num}}$ means the influence quantity of the given subject during inference.}
  \label{ablation_2}
\end{table}

We then ablate on the Top-K raito $\gamma_{\text{scope}}$ in Tab. \ref{ablation_1}: (1) within a proper range (experimentally, $\gamma_{\text{scope}} \in [0.2, 0.4]$) the results are quite robust; (2) the maximum normalization in Eq. \ref{max_norm} is important for the unity of high similarity and controllability, since different regions in the selected parts have different subject relevance and should be set to different weights. (3) Too small or too large influence scope will degrade similarity or controllability, respectively. These conclusions are validated by the visualization in Fig. \ref{topk_region}.

\textbf{Effectiveness of \emph{adaptive scoring module}.} As shown in Tab. \ref{ablation_2}, (1) We first compare with the simple use of all image features (ID-2), which results in degradation of both similarity and controllability, proving the importance of providing accurate and smooth influence quantity along with the coarse-to-fine diffusion generation process; (2) We then ablate on the module design (ID-3,4,5, ID-5), finding that using image score only results in worse performance. The reason is that the generation features are noisy at the beginning, resulting in an inaccurate score prediction. Therefore, we propose a step-scheduler to adaptively fuse text and image scores, leading to the best performance; (3) Finally, the choice of influence quantity $\gamma_{\text{num}}$ is ablated in ID-6 \& 7.
\section{Conclusion}
\label{sec:conclusion}

In this paper, we present a novel customization paradigm \emph{RealCustom} that, for the first time, disentangles similarity of given subjects from controllability of given text by precisely limiting subject influence to relevant parts, which gradually narrowing the real word from its general connotation to the specific subject in a novel ``train-inference" framework: the \emph{adaptive scoring module} learns to adaptively modulate influence quantity during training; (2) the \emph{adaptive mask guidance strategy} iteratively updates the influence scope and influence quantity of given subjects during inference. Extensive experiments demonstrate that RealCustom achieves the unity of high-quality similarity and controllability in the real-time open-domain scenario.


{
    \small
    \bibliographystyle{ieeenat_fullname}
    \bibliography{main}
}

\newpage

\section{Supplementary}
\label{sec:supplementary_main}

\subsection{More Qualitative Comparison}

\begin{figure*}
  \centering
  \includegraphics[width=1.0\linewidth]{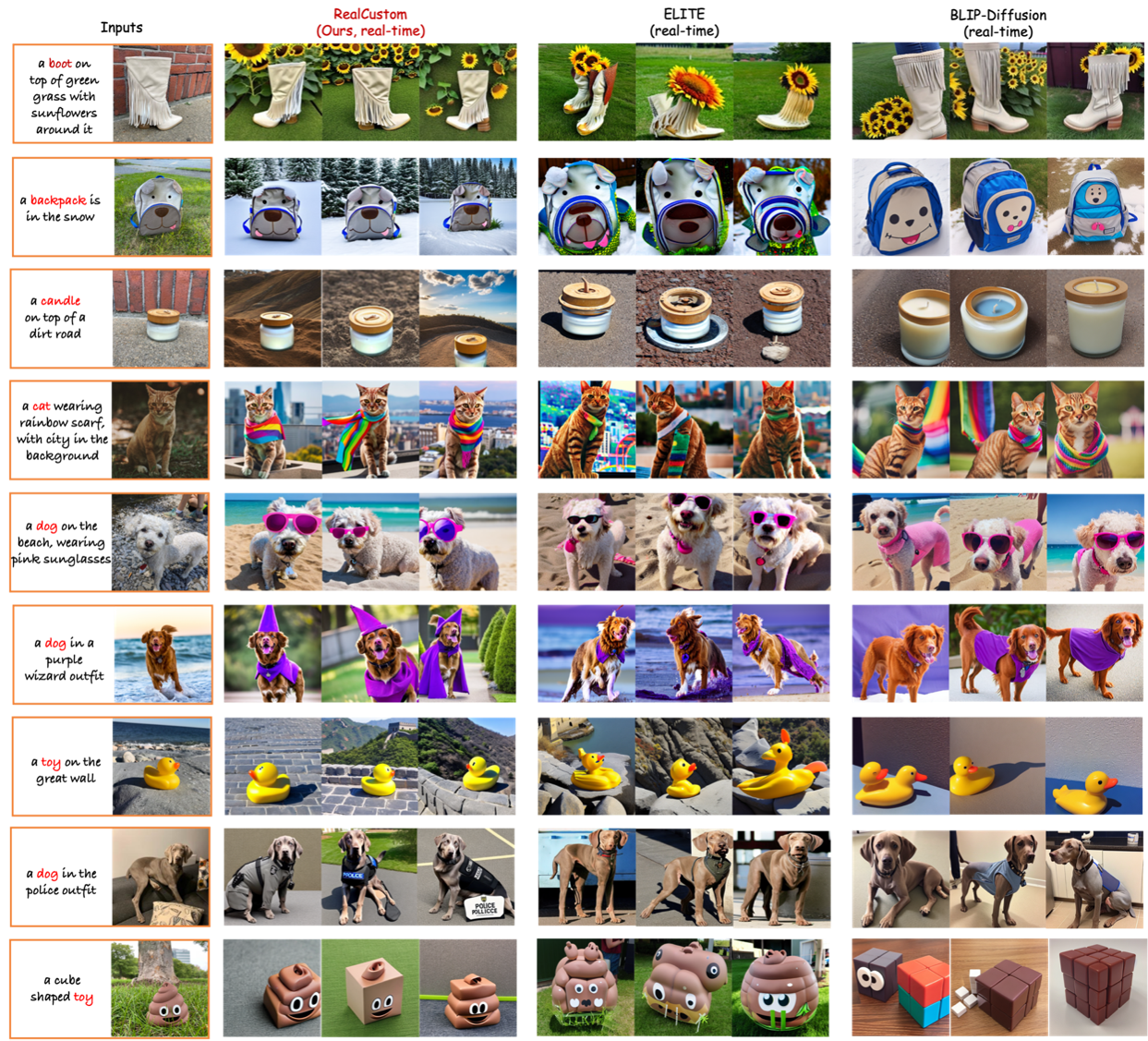}
  \caption{Qualitative comparison between our proposed \emph{RealCustom} and recent state-of-the-art methods of previous \emph{pseudo-word} paradigm in the \textcolor{blue}{\textbf{real-time}} customization scenario. We could conclude that (1) compared with existing state-of-the-arts, \emph{RealCustom} shows much \textcolor{blue}{\textbf{better similarity}} with the given subjects and \textcolor{blue}{\textbf{better controllability}} with the given text \textcolor{blue}{\textbf{at the same time}}, \eg, in the $7^{\text{th}}$ row, the toy generated by \emph{RealCustom} exactly on the Great Wall while existing works fail to adhere to the given text. Meanwhile, the toy generated by \emph{RealCustom} exactly mimics all details of the given one while existing works fail to preserve them. (2) \emph{RealCustom} generates customization images with much \textcolor{blue}{\textbf{better quality}}, \ie, better aesthetic scores, \eg, the snow scene in the second row, the dirt road scene in the third row, \etc. The conclusion adheres to our significant improvement (223.5\% improvement) on ImageReward \cite{xu2023imagereward} in the main paper since ImageReward evaluates both controllability and image quality. (3) \emph{RealCustom} shows \textcolor{blue}{\textbf{better generalization in open domain}}, \ie, for \emph{any given subjects}, \emph{RealCustom} could generate realistic images that consistently adhere to the given text for the given subjects in real-time, including the common subject like dogs (\eg, $5^{th}, 6^{th}$ rows) and rare subjects like the unique backpack (\ie, $1^{st}$ row), while existing state-of-the-arts works poorly on the rare subjects like the backpack in the first row, the special toy in the last row, \etc.}
  \label{Comparison}
\end{figure*}

As shown in Fig. \ref{Comparison}, we provide more qualitative comparison between our proposed \emph{RealCustom} and recent state-of-the-art methods of previous \emph{pseudo-word} paradigm in the \textcolor{blue}{\textbf{real-time}} customization scenario. Compared with existing state-of-the-arts, we could draw the following conclusions: 
(1) \textcolor{blue}{\textbf{better similarity}} with the given subjects and \textcolor{blue}{\textbf{better controllability}} with the given text \textcolor{blue}{\textbf{at the same time}}, \eg, in the $7^{\text{th}}$ row, the toy generated by \emph{RealCustom} exactly on the Great Wall while existing works fail to adhere to the given text. Meanwhile, the toy generated by \emph{RealCustom} exactly mimics all details of the given one while existing works fail to preserve them. 
(2) \textcolor{blue}{\textbf{better image quality}}, \ie, with better aesthetic scores, \eg, the snow scene in the second row, the dirt road scene in the third row, \etc. The conclusion adheres to our significant improvement (223.5\% improvement) on ImageReward \cite{xu2023imagereward} in the main paper since ImageReward evaluates both controllability and image quality. 
(3) \textcolor{blue}{\textbf{better generalization in open domain}}, \ie, for \emph{any given subjects}, \emph{RealCustom} could generate realistic images that consistently adhere to the given text for the given subjects in real-time, including the common subject like dogs (\eg, $5^{th}, 6^{th}$ rows) and rare subjects like the unique backpack (\ie, $1^{st}$ row), while existing state-of-the-arts works poorly on the rare subjects like the backpack in the first row, the special toy in the last row, \etc. The reason lies that for the very first time, our proposed \emph{RealCustom} progressively narrows a real text word from its initial general connotation into the unique subject, which completely get rid of the necessary corresponding between given subjects and learned pseudo-words, and therefore is no longer confined to be trained on object-datasets with limited categories.

\subsection{More Visualization}

\begin{figure*}
  \centering
  \includegraphics[width=1.0\linewidth]{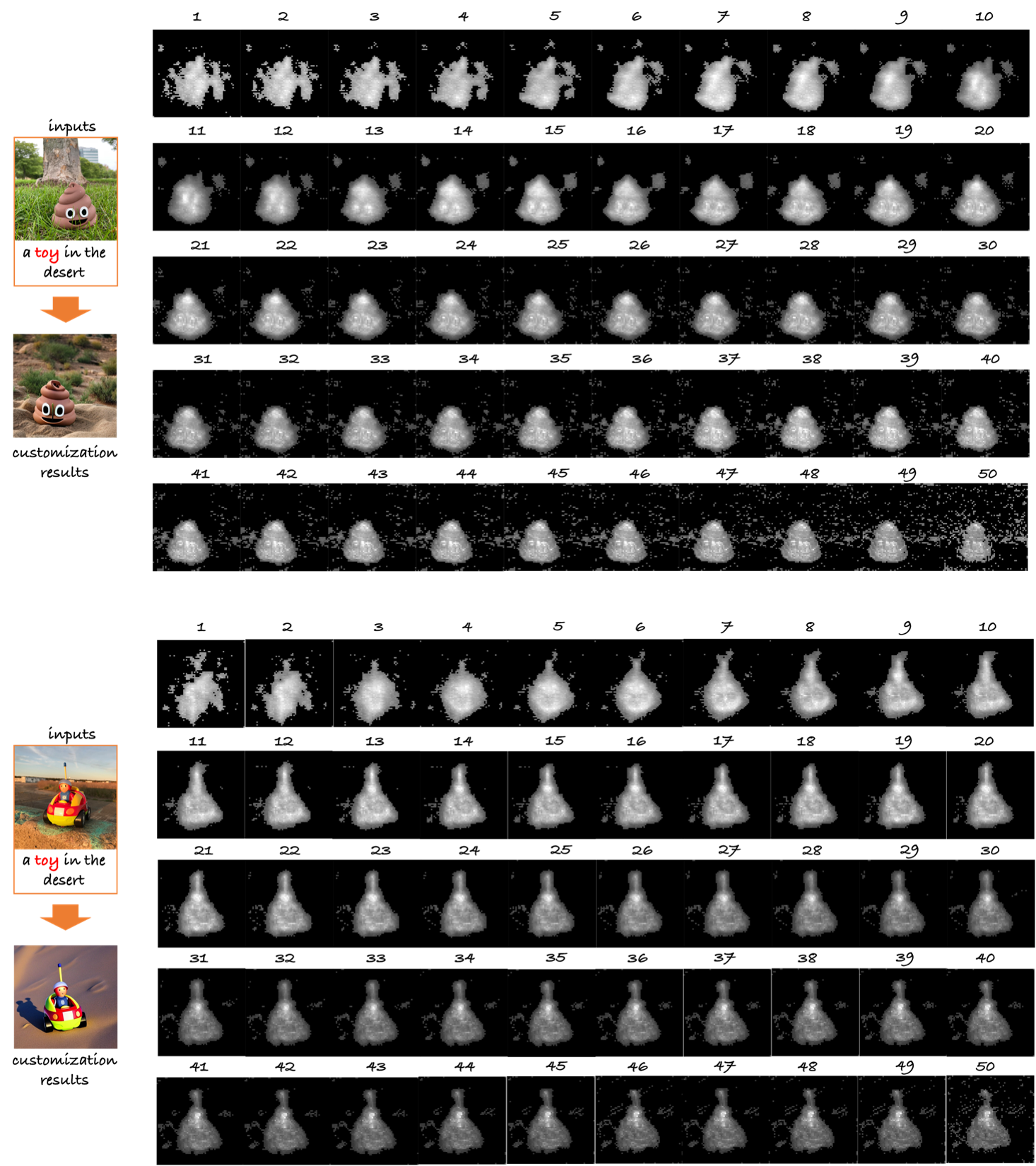}
  \caption{Illustration of gradually narrowing the real words into the given subjects.  Here we provide two customization cases that with the same given text ``a toy in the desert" and two different given subjects.  The real text word used for narrowing is ``toy".  The mask is visualized by the Top-25\% highest attention score regions of the real text word ``toy".  We visualize all the masks in the total 50 DDIM sampling steps, which are shown on the left.  We could observe that the mask of the ``toy" gradually being smoothly and accurately narrowed into the specific given subject.  Meanwhile, even in these subject-relevant parts (Top-25\% highest attention score regions of the real text word ``toy" in these cases), their relevance is also different, \eg, the more important parts like the eyes of the first subject are given higher weight (brighter in the mask). }
  \label{visualizae_1}
\end{figure*}

\begin{figure*}
  \centering
  \includegraphics[width=1.0\linewidth]{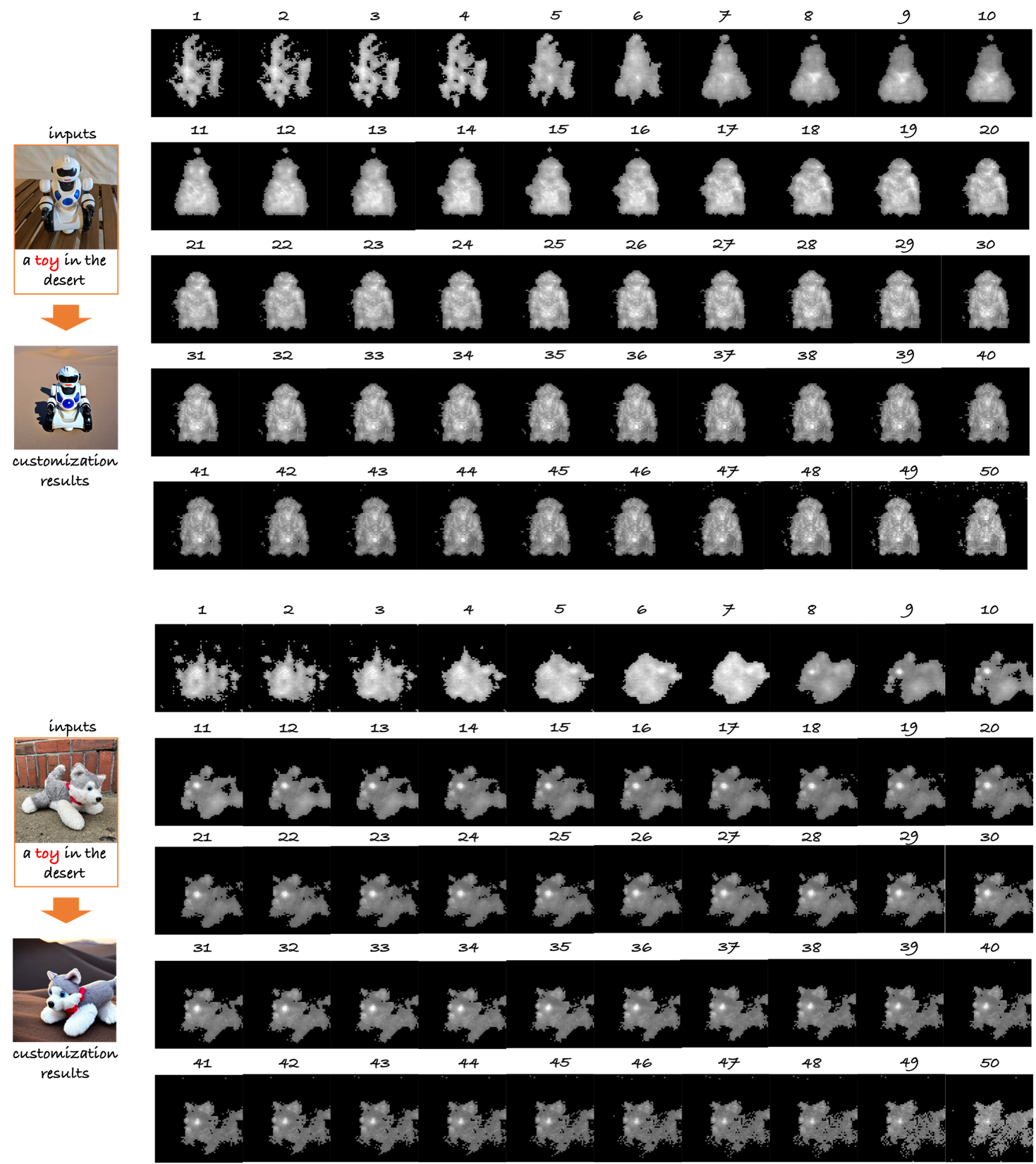}
  \caption{Illustration of gradually narrowing the real words into the given subjects.  Here we provide two customization cases that with the same given text ``a toy in the desert" and two different given subjects.  The real text word used for narrowing is ``toy".  The mask is visualized by the Top-25\% highest attention score regions of the real text word ``toy".  We visualize all the masks in the total 50 DDIM sampling steps, which are shown on the left.  We could observe that the mask of the ``toy" gradually being smoothly and accurately narrowed into the specific given subject.  Meanwhile, even in these subject-relevant parts (Top-25\% highest attention score regions of the real text word ``toy" in these cases), their relevance is also different, \eg, the more important parts like the eyes of the second subject are given higher weight (brighter in the mask).}
  \label{visualizae_2}
\end{figure*}

We provide more comprehensive visualization of the narrowing down process of the real word of our proposed \emph{RealCustom} in Fig. \ref{visualizae_1} and Fig. \ref{visualizae_2}. Here, we provide four customization cases that with the same given text ``a toy in the desert" and four different given subjects.  The real text word used for narrowing is ``toy".  The mask is visualized by the Top-25\% highest attention score regions of the real text word ``toy".  We visualize all the masks in the total 50 DDIM sampling steps.  We could observe that the mask of the ``toy" gradually being smoothly and accurately narrowed into the specific given subject.  Meanwhile, even in these subject-relevant parts (Top-25\% highest attention score regions of the real text word ``toy" in these cases), their relevance is also different, \eg, in Fig. \ref{visualizae_1}, the more important parts like the eyes of the first subject are given higher weight (brighter in the mask), in Fig. \ref{visualizae_2}, the more important parts like the eyes of the second subject are given higher weight.

\subsection{Impact of Different Real Word}

\begin{figure*}
  \centering
  \includegraphics[width=1.0\linewidth]{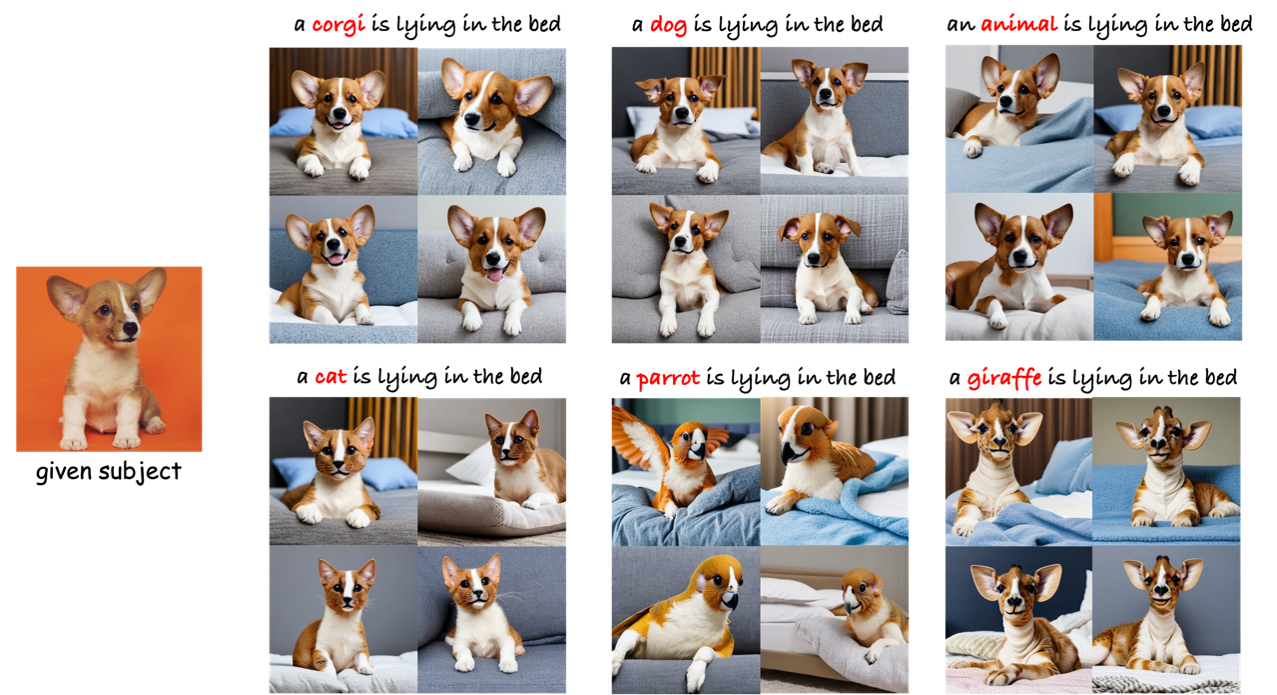}
  \caption{The customization results in using different real text words. The real text word narrowed down for customization is highlighted in red. We could draw the following conclusions: (1) The customization results of our proposed \emph{RealCustom} are \textcolor{blue}{\textbf{quite robust}}, \ie, no matter we use how coarse-grained text word to represent the given subject, the generated subject in the customization results are always almost identical to the given subjects. For example, in the upper three rows, when we use ``corgi", ``dog" or ``animal" to customize the given subject, the results all consistently adhere to the given subject. This phenomenon also validates the generalization and robustness of our proposed new paradigm \emph{RealCustom}. (2) When using \emph{completely different word} to represent the given subject, \eg, use ``parrot" to represent a corgi, our proposed \emph{RealCustom} opens a door for a new application, \ie, \textcolor{blue}{\textbf{novel concept creation}}. That is, \emph{RealCustom} will try to combine these two concepts and create a new one, \eg, generating a parrot with the appearance and character of the given brown corgi, as shown in the below three rows. This application will be very valuable for designing new characters in movies or games, \etc.}
  \label{word}
\end{figure*}

The customization results in using different real text words are shown in Fig. \ref{word}. The real text word narrowed down for customization is highlighted in red. We could draw the following conclusions: (1) The customization results of our proposed \emph{RealCustom} are \textcolor{blue}{\textbf{quite robust}}, \ie, no matter we use how coarse-grained text word to represent the given subject, the generated subject in the customization results are always almost identical to the given subjects. For example, in the upper three rows, when we use ``corgi", ``dog" or ``animal" to customize the given subject, the results all consistently adhere to the given subject. This phenomenon also validates the generalization and robustness of our proposed new paradigm \emph{RealCustom}. (2) When using \emph{completely different word} to represent the given subject, \eg, use ``parrot" to represent a corgi, our proposed \emph{RealCustom} opens a door for a new application, \ie, \textcolor{blue}{\textbf{novel concept creation}}. That is, \emph{RealCustom} will try to combine these two concepts and create a new one, \eg, generating a parrot with the appearance and character of the given brown corgi, as shown in the below three rows. This application will be very valuable for designing new characters in movies or games, \etc.


\end{document}